# Applications of natural language processing in aviation safety: A review and qualitative analysis

Aziida Nanyonga[1], Keith Joiner[2], Ugur Turhan[3] and Graham Wild[4]
*UNSW, Canberra, ACT, 2612, Australia*

**This study explores the use of Natural Language Processing (NLP) in aviation safety, focusing on machine learning algorithms designed to enhance safety measures. There are currently (May 2024), 34 Scopus results from the keyword search "natural language processing" and "aviation safety." Analyzing these studies allows us to uncover trends in the methodologies, findings, and implications of NLP in aviation. Both qualitative and quantitative tools have been used to investigate the current state of literature on NLP for aviation safety. The qualitative analysis summarizes the research motivations, objectives, and outcomes, showing how NLP can be utilized to help identify critical safety issues and improve aviation safety. This study also identifies research gaps and suggests areas for future exploration, providing practical recommendations for the aviation industry. We discuss challenges in implementing NLP in aviation safety, such as the need for large, annotated datasets, and the difficulty in interpreting complex models. We propose solutions like active learning for data annotation and explainable AI for model interpretation. Case studies demonstrate the successful application of NLP in improving aviation safety, highlighting its potential to make aviation safer and more efficient.**

## I. Nomenclature

| | |
|---|---|
| *AI* | = Artificial Intelligence |
| *ASN* | = Aviation Safety Network |
| *ASRS* | = Aviation Safety Reporting System |
| *BERT* | = Bidirectional Encoder Representations from Transformers |
| *DL* | = Deep Learning |
| *ICAO* | = International Civil Aviation Organization |
| *LDA* | = Latent Dirichlet Allocation |
| *LSA* | = Latent Semantic Analysis |
| *ML* | = Machine learning |
| *NLP* | = Natural language processing |
| *NLTK* | = Natural Language Toolkit |
| *NTSB* | = National Transportation Safety Board |
| *SHAP* | = SHapley Additive exPlanations |
| *SVD* | = Singular Value Decomposition |
| *Textalyser* | = Text analysis software |
| *VOSviewer* | = Visualization of similarities software |
| *WordItOut* | = Word cloud generator |

[1] PhD Candidate, School of Engineering and Technology.
[2] Senior Lecturer, School of Engineering and Technology, and AIAA Member.
[3] Senior Lecturer, School of Science.
[4] Senior Lecturer, School of Science, and AIAA Member.

## II. Introduction

Aviation safety is critical for the global aviation industry, where the primary goal is to minimize the number and severity of safety occurrences, either accidents or incidents, that may ultimately result in a loss of life and direct economic costs (hull losses etc.) [1]. Traditional methods of safety analysis have largely relied on manual inspection and categorization of incident reports, a process that is time-consuming and at risk of human error. In recent years, the advent of natural language processing (NLP) and machine learning (ML) has provided new opportunities to enhance the analysis of aviation safety data [2, 3] offering more efficient and accurate ways to uncover insights from large volumes of text-based reports [4, 5].

### A. Natural Language Processing

Natural language processing, a subfield of artificial intelligence (AI), focuses on the interaction between computers and human language. It enables the automatic extraction and processing of information from unstructured text, making it particularly suitable for analyzing aviation safety reports, which are often narrative-based and unstructured. ML algorithms, when combined with NLP techniques, can identify patterns, classify text, and predict outcomes based on the textual data available. These capabilities have significant implications for improving aviation safety by providing deeper insights into incident causation, operational inefficiencies, and emerging safety trends.

Various studies have demonstrated the application of NLP and ML in aviation safety. For instance, a study by Rose and colleagues [6] employed Singular Value Decomposition (SVD) to analyze dangerous goods incidents reported in the Aviation Safety Reporting System (ASRS), uncovering major topics related to such incidents. Similarly, Rankin et al. (2015) utilized Latent Semantic Analysis (LSA) to map primary causal factors in self-reported safety narratives, achieving notable accuracy in categorizing these narratives compared to expert huma results. Another significant work by Kuhn applied Structural Topic Modeling (STM) to ASRS and National Transportation Safety Board (NTSB) reports, highlighting the strengths of STM in identifying themes within technical datasets [7].

Advances in deep learning have further enhanced the potential of NLP in aviation safety. A study by Kierszbaum & Lapasset et al; [8] demonstrated the use of the Bidirectional Encoder Representations from Transformers (BERT) model for question answering on ASRS reports, achieving a 70% accuracy rate in extracting relevant information from free-text narratives. Additionally, Dong et al. [9] proposed a deep learning approach using Long Short-Term Memory (LSTM) networks to identify causal factors in incident reports, showing significant improvements over traditional methods.

### B. Air Transport System Safety

NASA's system safety program is the application of engineering and management principles, criteria, and techniques to achieve acceptable mishap risk within the constraints of operational effectiveness and suitability, time, and cost throughout all phases of the system life cycle [10]. It is an integral part of the interdisciplinary approach of systems engineering and its pursuit of systems that meet stakeholder expectations [10]. The methods of System Safety are diverse and are driven by many factors, including the high cost of testing, increasing system complexity, the development of systems that operate at the edge of engineering capability, and the use of unproven technology [10].

Predictive safety for the air transport system is a part of the System-Wide Safety (SWS) project [11]. The transformation of aviation to make flights more efficient and accessible, even as the global demand for air transportation services steadily increases, will require new research tools, innovative technologies, and operational methods [11]. Enabling this to be done safely at every step is the goal of the SWS project [11]. Predictive safety management in aviation involves the use of predictive (forecasting) and causal modelling methods to identify potential and possible hazards in the future, as well as their causal factors which can help define timely and efficient mitigation measures to prevent or restrain emerging hazards turning into adverse events [12] . This methodology is used in various aviation sectors (air navigation services, airport operations, airline operations) for planning purposes [13]. It is a crucial mechanism to maintain and continuously improve safety levels in aviation organizations [13].

### C. Motivation and Significance

The motivation behind integrating NLP into aviation safety is multi-faceted. First, it addresses the need for more accurate, proactive, and efficient data processing techniques that can handle the growing volume of safety reports. Second, it enhances the accuracy of safety analysis by reducing the subjectivity and potential biases inherent in manual categorization [14]. Third, it facilitates the early detection of safety trends and the identification of latent issues that may not be immediately apparent through conventional methods [15].

More importantly, it is a steppingstone towards qualitative predictive safety. That is, the development of language models and qualitative analytics using AI/ML facilitates not only contemporary quantitative predictive safety but enables sources of data like voice comms and ACARS messages, etc., to be used as sources of data that can be tracked over time looking for precursors to potential future safety occurrences. While the concept of “big brother” is not the goal, looking at an accident such as Colgan Air Flight 3407 [16], precursors to safety may very well be embedded in audio data. Similarly, text in check-and-training reports, or other continuous documentation may uncover issues. However, in the context of NLP for aviation safety occurrence reports, there are precursors in prior incidents that were near misses, which can be learnt from (as is currently done through the safety culture in aviation), and when similar situations are encountered during operations, a preventative measure may be taken to prevent an accident from happening, an artificial memetic algorithm.

### D. Aim and Research Objectives

The overarching aim of this study is to critically evaluate the role of NLP in enhancing aviation safety through a systematic review and qualitative analysis of existing literature. By examining a selection of studies indexed in Scopus, this research seeks to synthesize the current state of knowledge regarding the application of NLP in aviation safety, identify gaps in the literature, and propose directions for future research. The associated research questions are:

1) What are the machine learning algorithms used for NLP in aviation safety, and how do they contribute to the field?
2) How does NLP facilitate the identification and analysis of critical safety issues within the aviation industry?
3) What challenges are associated with implementing NLP in aviation safety, and what solutions have been proposed or implemented to address these challenges?
4) In what ways can NLP contribute to the advancement of predictive safety management within the aviation sector?
5) What are the implications of NLP applications in aviation safety for industry stakeholders, and how can these insights inform policy-making and best practices?

## III.Methodology

The methodology employed in this study involved a detailed examination of 34 elected papers focusing on natural language processing (NLP) applications in aviation safety. The selection process was guided by a systematic literature review titled "A Systematic Review of Machine Learning Analytic Methods for Aviation Accident Research," currently under review for publication [17]. This review identified relevant papers through comprehensive searches on Scopus and backward reference searches via Google Scholar, resulting in the identification of 87 papers.

From this pool, the first seventeen papers were chosen for inclusion in this study based on their relevance to NLP applications in aviation safety. These papers were further filtered based on their utilization of data sources from authoritative aviation safety organizations such as the Aviation Safety Reporting System (ASRS), International Civil Aviation Organization (ICAO), National Transportation Safety Board (NTSB), Australian Transport Safety Bureau (ATSB), and Aviation Safety Network (ASN). Papers selected for inclusion were published up to the year 2022, consistent with the timeframe of the systematic literature review. Also, to ensure the timeliness of the findings, papers from 2023 and 2024 were also included, resulting in 34 papers reviewed.

### A. Data Collection

Detailed information from each paper, including the year of publication, title, data source, sample size, aviation application, machine learning algorithms employed, comments, research questions/objectives, study design, outcomes/conclusions, and keywords, was systematically extracted.

### B. Statistical Analysis

Word clouds were generated using free online tools such as WordItOut and Textalyser to visualize frequently occurring terms in the selected papers. This analysis provided insights into the prominent themes and topics addressed in the literature. Additionally, keyword counts were performed to identify prevalent themes and topics across the selected papers. This quantitative analysis helped in understanding the distribution and significance of key terms within the literature. VOSviewer was utilized to generate maps and networks based on the research objectives and abstracts from all the papers, facilitating a comprehensive visualization of the relationships and connections between key concepts and topics in the research [18, 19].

### C. Study Design Considerations

In the study design considerations, the research questions and objectives of each paper were systematically analyzed to understand the overarching goals and motivations behind the studies. Additionally, the methodologies employed in the selected papers were examined to identify common approaches and techniques used in NLP applications for aviation safety. The outcomes and conclusions drawn from each paper were carefully reviewed to extract insights and implications for future research and practice in the field.

## IV. Advancing Aviation Safety

The integration of NLP into aviation safety research is driven by several compelling factors. The aviation industry generates much safety data through various reporting systems such as the Aviation Safety Reporting System (ASRS) and the National Transportation Safety Board (NTSB) reports. These reports, often rich in narrative content, provide detailed accounts of incidents and accidents. However, the unstructured nature of pure text narratives poses a significant challenge for traditional data analysis methods, which are typically designed for structured data. NLP offers a powerful solution by enabling the automated extraction and processing of meaningful information from large volumes of text, thereby enhancing the analysis of safety data.

### A. Efficiency and Scalability

One of the primary motivations for employing NLP in aviation safety is the need for more efficient and scalable data processing techniques. Manual analysis of safety reports is not only labor-intensive but also time-consuming, limiting the ability to promptly address emerging safety concerns. NLP techniques, such as topic modelling, clustering, and semantic analysis, can process vast amounts of text data quickly and accurately. For instance, a study by Walton et al. [20] demonstrated how SVD could be used to analyze dangerous goods incidents from the ASRS database, efficiently uncovering key topics and trends over ten years.

### B. Accuracy and Objectivity

Traditional methods of safety data analysis often rely on subjective judgment and manual categorization, which can introduce biases and inconsistencies. NLP algorithms, by contrast, offer a more objective approach to text analysis. Machine learning models such as LSA and BERT have been shown to achieve high levels of accuracy in categorizing and extracting information from safety narratives. For example, Robinson [21] utilized LSA to map primary causal factors in safety reports, demonstrating an unsupervised categorization accuracy of 44% within an existing taxonomy. Similarly, Kierszbaum & Lapasset et al. [8] applied BERT for question answering on ASRS reports, achieving roughly 70% correct answers, other studies [22-27] also applied BERT to their studies thereby highlighting the model's capability to handle the complexity of natural language narratives.

### C. Early Detection and Proactive Safety Management

Another significant motivation for applying NLP in aviation safety is its potential to facilitate the early detection of safety trends and latent issues. By analyzing large datasets of safety reports, NLP can identify patterns and correlations that might not be immediately apparent through manual analysis. This proactive approach allows for the identification of emerging risks and the development of preventive measures before issues escalate into more severe incidents. For instance, studies such as [2, 3, 9, 28-31] applied Deep Learning (DL) algorithms to aviation safety data, identifying themes and correlations within safety narratives that could inform future safety studies and interventions.

### D. Enhancing Human Factors Analysis

Human factors play a crucial role in aviation safety, with many incidents and accidents attributed to human error. NLP techniques can significantly enhance the analysis of human factors by extracting relevant information from narrative reports and identifying underlying causes and contributing factors [24]. Dong et al. proposed a deep learning approach using LSTM networks to identify causal factors in incident reports, focusing on human error and other contributory elements. Their model demonstrated significant improvements in accuracy and efficiency compared to traditional analysis methods [9].

### E. Contribution to Knowledge and Best Practices

The application of NLP in aviation safety improves immediate safety outcomes and contributes to broader knowledge in the field. By providing a deeper understanding of safety issues and their underlying causes, NLP-driven research informs best practices and policymaking. Studies such [6, 32, 33] exemplify how NLP can generate valuable insights that enhance aviation safety protocols and decision-making processes.

## V. Navigating Aviation Safety

The use of NLP in aviation safety has been explored across various dimensions, encompassing a range of ML algorithms and techniques tailored to different safety datasets. This chapter reviews the specific applications of NLP in aviation safety, drawing on insights from 34 selected studies. We examine the methodologies used, research objectives, study designs, and key outcomes to understand how NLP has been used to enhance aviation safety.

### A. Methodologies and Machine Learning Algorithms

Different NLP methodologies and machine learning algorithms have been applied to aviation safety data, each with its unique strengths and applications. The following sections detail these methodologies and their applications:

1) SVD: Walton and Marion utilized SVD to analyze dangerous goods incidents reported in the ASRS. SVD, a dimensionality reduction technique, was employed to identify key topics and trends within the safety narratives. This approach enabled the extraction of meaningful patterns from a large and complex dataset, demonstrating the utility of SVD in making sense of unstructured text data [20].

2) LSA: Robinson et al. applied LSA to map primary causal factors in self-reported safety narratives. LSA is a statistical method for extracting and representing the contextual usage meaning of words by analyzing relationships between a set of documents and the terms they contain. This technique proved effective in categorizing safety narratives and identifying primary causal factors, highlighting the potential of LSA in contextual and thematic evaluation of safety reports [21].

3) STM: Rose et al. (2022) employed STM to analyze safety data from the ASRS and NTSB. STM is a probabilistic model that identifies topics within a set of documents and can accommodate metadata to improve topic coherence. This study demonstrated the capability of STM to uncover latent themes within technical datasets, which can inform safety interventions and policymaking (Rose et al., 2022).

4) BERT and Deep Learning: Kierszbaum and Lapasset explored the use of the BERT model for answering questions on ASRS reports. BERT, a transformer-based model, excels in understanding the context of words in search queries and is particularly effective for tasks requiring a deep understanding of language. Their study showed that BERT could accurately extract relevant information from free-text narratives, indicating its potential for improving information retrieval in safety databases [8]. Dong et al. [9] developed a deep learning approach using LSTM networks to identify causal factors in incident reports. LSTM networks are well-suited for sequence prediction tasks and can effectively handle the complexity of narrative data. This approach enhanced the identification of primary and contributing factors, demonstrating the advantages of deep learning in the causal analysis of aviation incidents [9].

### B. Research Objectives and Study Designs

The research objectives and study designs across the reviewed studies varied, reflecting the diverse applications of NLP in aviation safety. Common objectives included improving the efficiency of data processing, enhancing the accuracy of safety analysis, and identifying latent safety issues. The following sections summarize the research questions and study designs employed in the selected studies.

1) Improving Data Processing Efficiency: Several studies aimed to develop frameworks and methodologies to process safety data more efficiently. For instance, Walton and Marion focused on establishing a data-driven framework using SVD to analyze dangerous goods incidents. This study highlighted the potential of NLP techniques to handle large datasets and uncover significant trends without extensive manual intervention [20].

2) Enhancing Accuracy of Safety Analysis: Other studies aimed to improve the accuracy and objectivity of safety analysis. Robinson et al. sought to evaluate the effectiveness of LSA in categorizing safety narratives and identifying primary causal factors. Their findings demonstrated that NLP could enhance the precision of safety data analysis, reducing the reliance on subjective judgment [21].

3) Identifying Latent Safety Issues: In his study, Kuhn aimed to identify latent topics and trends within aviation safety reports using STM. This approach revealed previously unreported connections and themes, underscoring the capability of NLP to detect emerging risks and inform proactive safety management [7].

### C. Key Outcomes and Implications

The reviewed studies collectively highlight the significant contributions of NLP to aviation safety. Key outcomes include improved data processing efficiency, enhanced accuracy in safety analysis, and the identification of latent safety issues. These outcomes have important implications for the aviation industry, suggesting that NLP can transform safety management practices by providing deeper insights into safety data.

For instance, the study by Kierszbaum and Lapasset [8] demonstrated that BERT could accurately extract relevant information from safety reports, suggesting its potential for improving the efficiency of information retrieval in safety databases. Similarly, the work by Dong et al. [9] showed that deep learning models could effectively identify causal factors in incident reports, providing valuable insights for preventive measures and safety interventions.

## VI. Quantitative Review

We used the *Satisfy* app (*https://statisty.app/)*, a publicly available online tool, for statistical analysis of the selected papers. Fig. 1 shows the Receiver Operating Characteristic (ROC) curve, illustrating the sensitivity versus specificity of the selected words, with an Area Under the Curve (AUC) of 0.879, indicating a good level of discrimination between relevant and irrelevant words in the 34 NLP papers. Fig. 2 presents a Sankey diagram illustrating the flow of data sources, aviation applications, and machine learning algorithms identified in the study. The ASRS (Aviation Safety Reporting System) was the most utilized data source, featuring in more than 20 studies (57% of the total), all aviation constituted 27 out of the 34 studies, followed by airline-specific studies with 6, and cargo-related studies with 1. Deep Learning (DL) and Latent Dirichlet Allocation (LDA) were the most frequently used algorithms, each appearing in 23% of the studies, with BERT used in 20% of the studies.

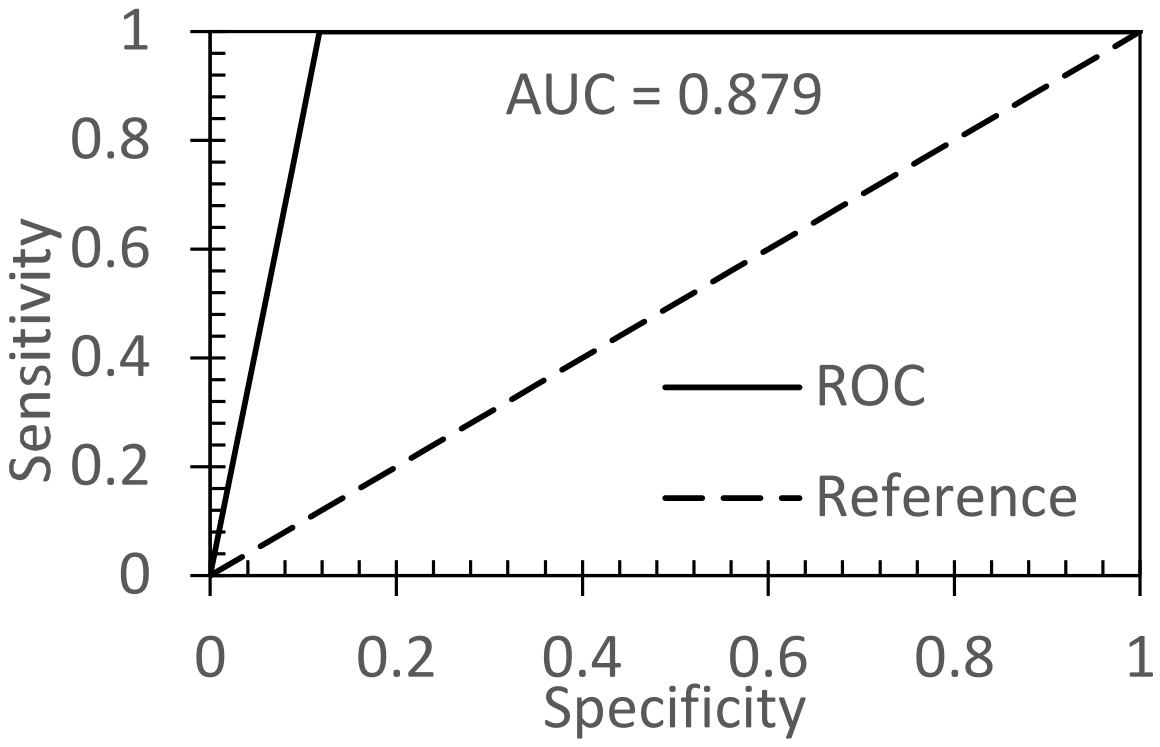

**Fig. 1. ROC Curve**

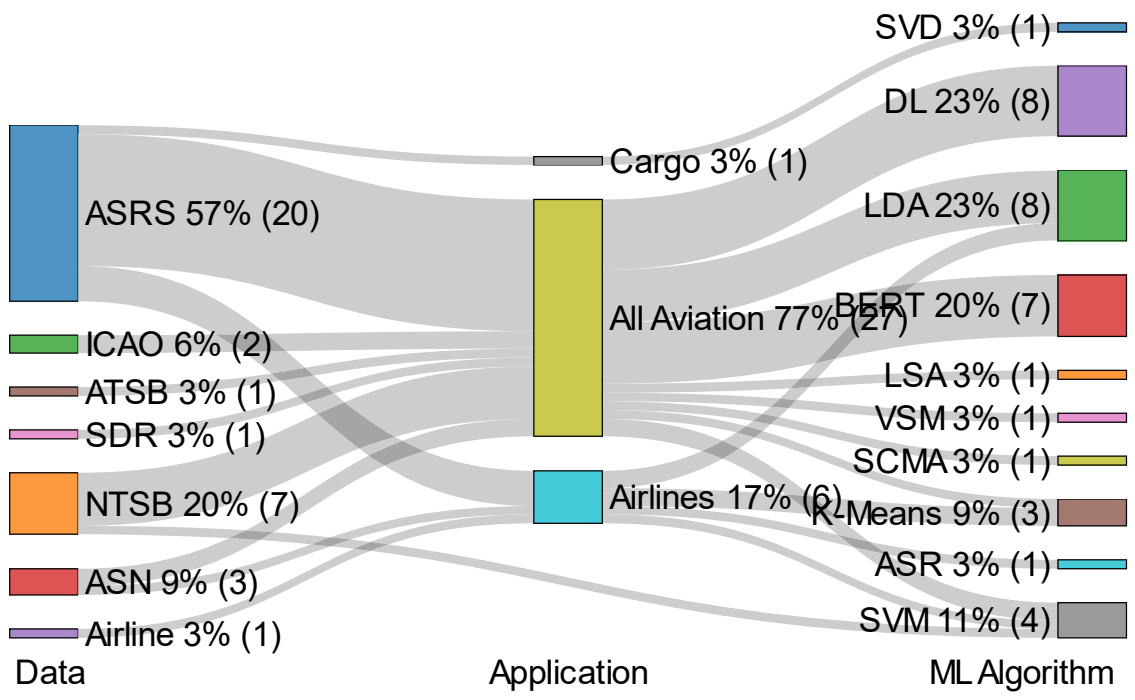

**Fig. 2. Sankey Diagram**

Fig. 3 shows the distribution of studies applying different ML algorithms across various aviation applications, highlighting the diverse application of ML techniques in enhancing aviation safety. Fig. 4 and Fig. 7, illustrates the most frequently used words across the studies, with "NLP" (Natural Language Processing) appearing most frequently with over 17 mentions, followed by "ASRS" (12 mentions), "Text mining" (9 mentions), and "Aviation safety" (8 mentions).

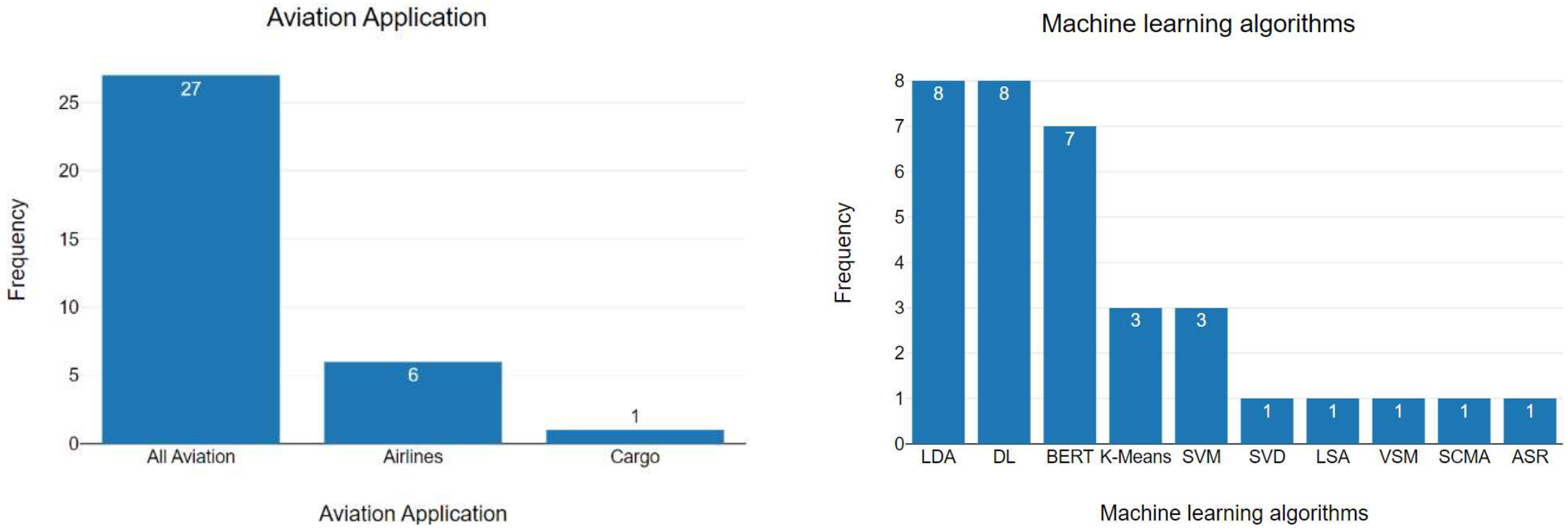

(A) (B)

**Fig. 34: The number of studies applying (A) different ML algorithms, across (B) various aviation applications.**

Next, we employed VOSviewer for bibliometric analysis of the keywords in the selected publications related to NLP and aviation safety. Fig. 5 presents the co-occurrence of keywords within the same publications. The size of the nodes represents the frequency of occurrence, while the curves between nodes indicate their co-occurrence; the heat map in Fig. 5 (b) also illustrates the relative size of the nodes as a hotter colour. Keywords such as “natural language processing”, “aviation safety”, and “text mining” were the most frequent, while terms like “fires” and “nasa” were less common. The shorter distance between nodes indicates a higher number of co-occurrences, suggesting strong thematic connections between frequently co-occuring keywords. Fig. 6 shows a Network visualizing the frequency of all keywords. The font size in the word cloud represents the frequency of occurrence, with terms like “natural language processing”, “aviation safety”, and “text mining” dominating, reflecting their central importance in the reviewed studies. These analyses highlight the prevalent themes and methodological trends in applying NLP and machine learning to aviation safety, emphasizing the field's complexity and the necessity for robust analytical approaches. Table 1 shows Aviation Applications across various ML algorithms while Table 2 summarizes different key findings identified from Aviation Safety Reports

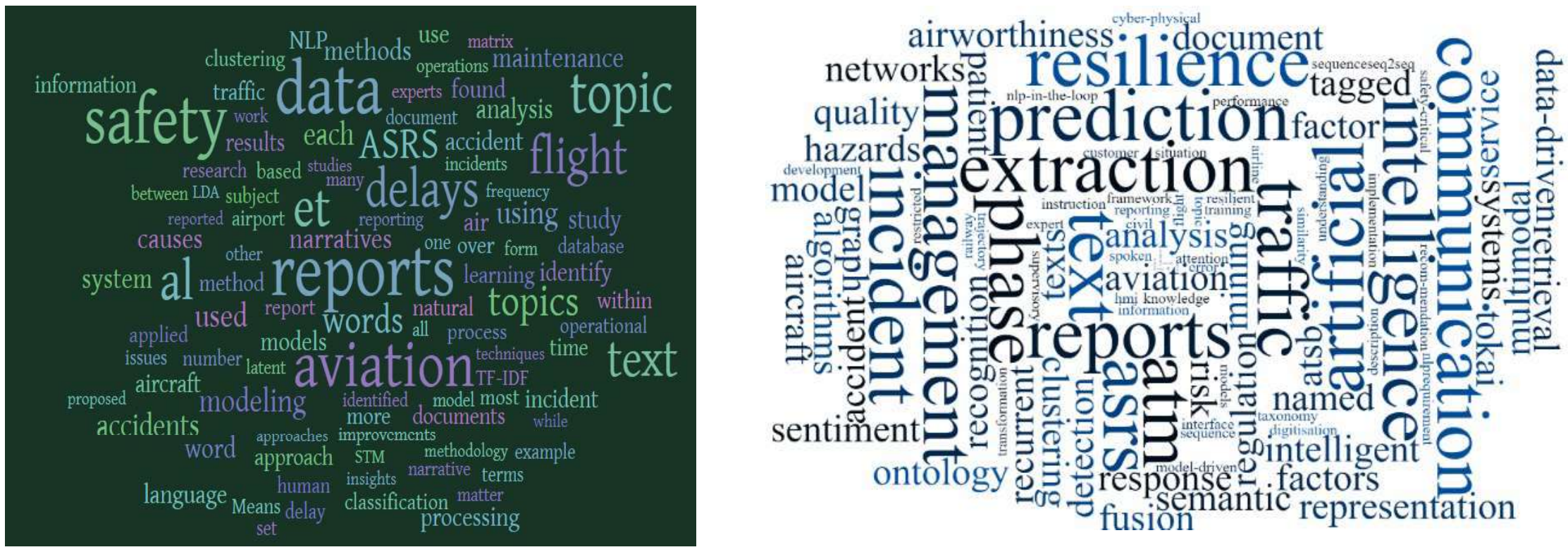

(A) (B)

**Fig. 5: Most frequently used words across the studies (A) objectives and (B) keywords**

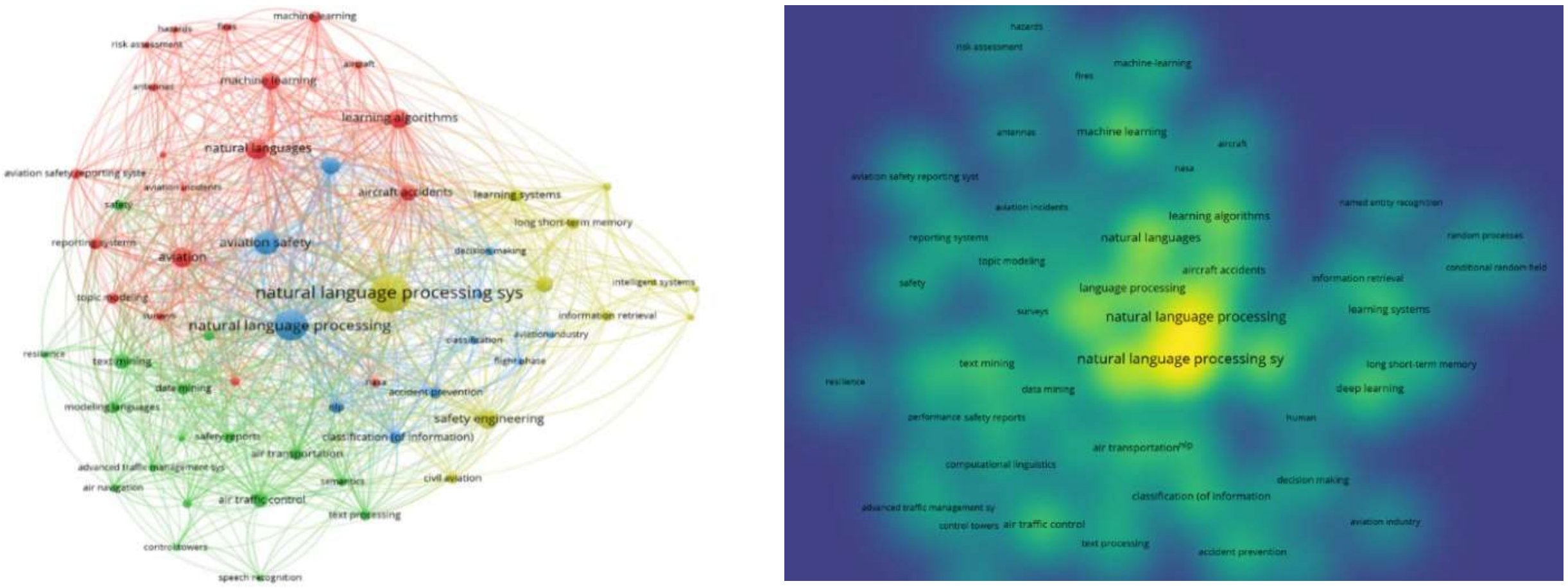

(A) (B)

**Fig. 6: Shows the co-occurrence of words (A) Network and (B) mapping within the same publications.**

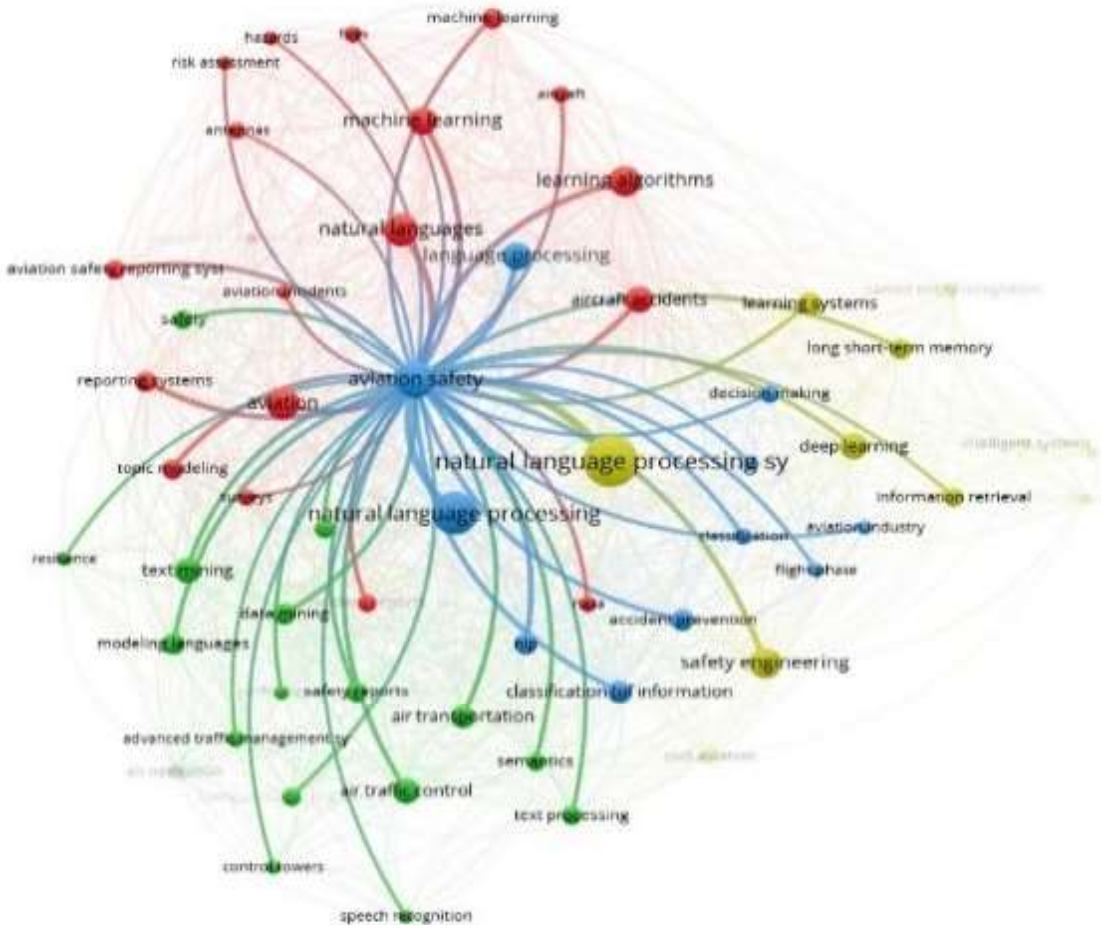

**Fig. 7. Network Visualization of Materials Keyword**

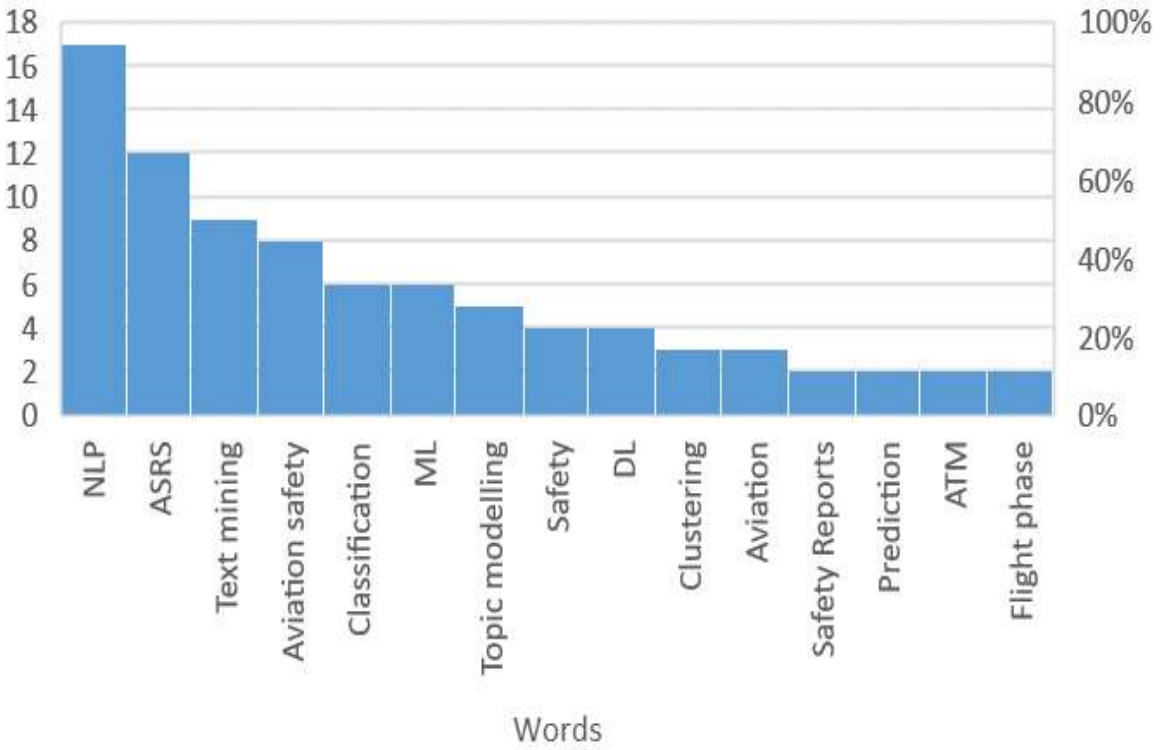

**Fig. 8. Word frequency in the 34 studies**

**Table 1: Distribution of ML algorithms and Aviation application**

| | SVD | LSA | LDA | BERT | DL | K-Means | SCMA | ASR | SVM | VSM | Total |
|---|---|---|---|---|---|---|---|---|---|---|---|
| Cargo | 1 | 0 | 0 | 0 | 0 | 0 | 0 | 0 | 0 | 0 | 1 |
| All Aviation | 0 | 1 | 7 | 7 | 8 | 1 | 1 | 0 | 1 | 1 | 27 |
| Airlines | 0 | 0 | 2 | 0 | 0 | 2 | 0 | 1 | 2 | 0 | 7 |
| Total | 1 | 1 | 9 | 7 | 8 | 3 | 1 | 1 | 3 | 1 | 35 |

**Table 2: ML Algorithms Applied to Aviation Safety Reports**

| Authors | Year | Data | Application | Algorithms |
|---|---|---|---|---|
| Walton & Marion [20] | 2020 | ASRS | Cargo | SVD |
| Robinson et al; [21] | 2015 | ASRS | All Aviation | LSA |
| Rose et al; [34] | 2022 | ASRS | Airlines | LDA |
| Kierszbaum & Lapasset [8] | 2020 | ASRS | All Aviation | BERT |
| Paradis et al; [35] | 2021 | ASRS | All Aviation | LDA |
| Chanen [28] | 2016 | ASRS | All Aviation | DL |
| Dong et al; [9] | 2021 | ASRS | All Aviation | DL |
| Agarwal et al; [22] | 2022 | NTSB | All Aviation | BERT |
| Nanyonga et al; [3] | 2021 | NTSB | All Aviation | DL |
| Rose et al; [6] | 2020 | ASRS | Airlines | K-Means |
| Perboli et al; [36] | 2021 | ICAO | All Aviation | VSM |
| Miyamoto et al; [33] | 2022 | ASRS | Airlines | K-Means |
| Nanyonga et al; [2] | 2023 | NTSB | All Aviation | DL |
| Robinson, [37] | 2019 | ASRS | Airlines | LDA |
| Zhao, et al; [38] | 2022 | NTSB | All Aviation | LDA |
| Abdhul et al; [39] | 2021 | ASRS | All Aviation | LDA |
| Kuhn, [7] | 2018 | ASRS | All Aviation | LDA |
| Tanguy et al; [40] | 2016 | ASRS | All Aviation | SVM |
| Madeira et al; [41] | 2021 | ASN | Airlines | SVM |
| Nanyonga et al; [3] | 2023 | NTSB | All Aviation | DL |
| Luo & Shi [42] | 2019 | ASRS | All Aviation | LDA |
| Kierszbaum et al; [23] | 2021 | ASRS | All Aviation | BERT |
| Switzer et al; [32] | 2011 | ASRS | All Aviation | SCMA |
| Buselli et al; [43] | 2022 | ICAO | All Aviation | LDA |
| Rose et al; [6] | 2020 | ASRS | All Aviation | K-Means |
| Abdullah et al; [44] | 2017 | Airline | Airlines | ASR |
| Nanyonga et al; [29] | 2023 | ATSB | All Aviation | DL |
| Persing [45] | 2009 | ASRS | All Aviation | SVM |
| Ono & Nakanishi [24] | 2023 | ASRS | All Aviation | BERT |
| Mbaye et al; [25] | 2023 | ASRS | All Aviation | BERT |
| Nanyonga & Wild [30] | 2023 | ASN; NTSB | All Aviation | DL |
| Nanyonga et al; [31] | 2023 | ASN | All Aviation | DL |
| Niraula et al; [26] | 2023 | SDR | All Aviation | BERT; SVM |
| Gao et al; [27] | 2024 | NTSB | All Aviation | BERT |

## VII. Discussion

### A. Key Trends and Methodologies

The synthesis of findings from the reviewed studies reveals several key trends and insights into the application of natural language processing (NLP) in aviation safety. Across the studies, common methodologies and machine learning algorithms were employed to analyze textual data from various aviation safety sources, including the Aviation Safety Reporting System (ASRS) and the National Transportation Safety Board (NTSB). The effectiveness of these methodologies varied based on the specific objectives of each study and the nature of the aviation applications examined.

*1. Topic Modeling Techniques*

Several studies utilized topic modelling techniques such as Latent Dirichlet Allocation (LDA) and Structural Topic Modeling (STM) to identify latent topics and themes within aviation safety narratives. These approaches enabled researchers to uncover underlying patterns and trends in safety reports, facilitating a deeper understanding of safety-related issues and contributing to proactive risk mitigation efforts.

*2. Deep Learning Models*

The application of deep learning models, including Long Short-Term Memory (LSTM) networks and Bidirectional LSTM (BLSTM), demonstrated promising results in extracting meaningful insights from aviation safety narratives. These models enhanced the accuracy of incident classification and causal factor identification, providing valuable support for safety analysis and decision-making processes.

*3. Semantic Analysis and Word Embeddings*

Studies leveraging semantic analysis techniques such as Latent Semantic Analysis (LSA) and word embeddings (e.g., word2vec) highlighted the ability to capture word-level meaning and semantic relationships within safety reports. By quantifying semantic similarities and associations, these approaches facilitated a more nuanced analysis of safety narratives and improved the interpretability of model outputs.

*4. Implications for the Industry*

The findings from the reviewed studies have significant implications for the aviation industry, particularly in enhancing safety management practices and accident prevention strategies. By leveraging NLP technologies and machine learning algorithms, aviation stakeholders can:

- Gain deeper insights into safety-related incidents and causal factors.
- Prioritize safety interventions based on identified trends and patterns.
- Improve the efficiency of safety reporting and analysis processes.
- Enhance predictive capabilities for identifying potential safety risks and hazards.
- Facilitate evidence-based decision-making and proactive risk mitigation efforts.

## B. Research Gaps and Future Directions

Despite significant advancements in applying natural language processing (NLP) to aviation safety, several research gaps remain that warrant further exploration. This chapter identifies these gaps and proposes potential areas for future research, along with specific questions and topics that should be addressed.

*1. Research Gaps*

One significant gap is the limited integration of multimodal data sources. Most current studies focus exclusively on textual data from safety reports, neglecting other valuable data types such as audio recordings, sensor data, and video footage. Integrating these diverse data sources could provide a more comprehensive understanding of safety incidents and help identify underlying causes that are not apparent from text alone.

Another gap is the underutilization of real-time data. Most research deals with historical data, with minimal emphasis on real-time data analysis. Real-time processing and analysis could enable proactive safety measures and immediate response strategies, significantly enhancing aviation safety management. The ability to detect and address safety issues as they occur, rather than retrospectively, could lead to more effective interventions and reduced incident rates. The current literature also shows an insufficient focus on human factors. Although some studies have explored the role of human factors in aviation safety, this area remains underdeveloped. There is a need for a more in-depth analysis of cognitive and behavioral aspects to understand their impact on safety incidents comprehensively. Factors such as fatigue, decision-making processes, and communication dynamics are critical to aviation safety but are not adequately addressed by existing NLP models.

Additionally, there is a lack of research encompassing diverse geographic and cultural contexts. Most studies are concentrated in specific regions, particularly the United States and Europe, with limited research on aviation safety in other parts of the world. Expanding research to include a variety of geographic and cultural contexts could uncover unique safety challenges and solutions applicable to different regions. This diversity would help develop more universally applicable safety protocols and interventions.

Another significant gap is the lack of standardized evaluation metrics. There is a considerable inconsistency in how the effectiveness of NLP models in aviation safety applications is measured, making it challenging to compare results across different studies and identify best practices. Standardizing these metrics would facilitate better benchmarking and improvement of NLP techniques in the field.

*2. Future Directions*

Future research should focus on the integration of multimodal data. Methods for combining textual, audio, video, and sensor data should be developed to provide a holistic view of safety incidents. Techniques such as multimodal deep learning and data fusion could be employed to achieve this integration, leading to more comprehensive and accurate safety analyses.

Real-time data analysis represents another promising area for future exploration. Developing frameworks and algorithms capable of handling streaming data from various sources in real-time can enable immediate safety interventions and improve situational awareness. Research should focus on creating robust real-time data processing systems that can detect and respond to safety issues as they occur, potentially preventing incidents before they escalate.

Comprehensive studies on human factors should be prioritized. Future research should delve deeper into cognitive and behavioral aspects of aviation safety, using NLP to analyze how these factors influence safety outcomes. Investigating the impact of human factors such as fatigue, stress, communication breakdowns, and decision-making processes could provide valuable insights into preventing human error-related incidents.

Expanding research to include diverse geographic and cultural contexts is also essential. By conducting studies in various regions and cultural settings, researchers can uncover unique safety challenges and develop solutions that are globally applicable. This approach would help create more effective and inclusive safety protocols and interventions.

Standardizing evaluation metrics for NLP models in aviation safety applications should be another focus. Establishing consistent benchmarks for assessing the performance of these models would facilitate better comparison of research outcomes and drive improvements in NLP techniques. Developing a set of agreed-upon metrics would enhance the reliability and validity of research findings in this field.

*3. Specific Questions and Topics for Future Research*

To address these research gaps, future studies should consider the following questions and topics:

1. How can multimodal data integration improve the accuracy and comprehensiveness of aviation safety analyses?
2. What frameworks and algorithms are most effective for real-time data processing in aviation safety?
3. What are the critical human factors influencing aviation safety, and how can they be effectively analyzed using NLP?
4. How do aviation safety challenges and solutions vary across different geographic and cultural contexts?
5. What standardized metrics should be developed to evaluate the effectiveness of NLP models in aviation safety applications?

## C. Practical Recommendations

*1. Integration of NLP into Safety Management Systems*

NLP can significantly enhance safety management systems by enabling the analysis of textual data from safety reports. By integrating NLP tools into existing systems, aviation stakeholders can extract valuable insights from narrative data, facilitating more informed decision-making and proactive safety measures. For instance, NLP algorithms can analyze incident reports, identify recurring patterns, and prioritize corrective actions based on risk assessment.

*2. Overcoming Challenges in Data Annotation and Model Interpretation*

An important challenge in implementing NLP solutions is annotating data required for model training. To address this, the industry can develop standardized annotation guidelines and leverage automated annotation tools to streamline the process. Additionally, incorporating expert review and feedback mechanisms can ensure the accuracy and relevance of annotated data, thereby enhancing the quality of NLP models. Ensuring transparency and interpretability of NLP models is essential for gaining stakeholders' trust and facilitating model adoption. To achieve this, aviation organizations should employ interpretable machine learning methods and visualization techniques to make model outputs understandable to safety analysts. Techniques such as SHAP and LIME can provide insights into the underlying mechanisms of NLP models, aiding in model interpretation and decision-making.

*3. Guidance on Model Integration, Deployment, and Capacity Building*

Effective integration and deployment of NLP models require careful consideration of organizational requirements and constraints. Aviation stakeholders should develop robust model integration frameworks and establish clear guidelines for model deployment. This includes defining model performance metrics, establishing protocols for model updates and maintenance, and ensuring compliance with regulatory standards and industry best practices. Continued investment in research and development is crucial for advancing NLP applications in aviation safety. Industry stakeholders should collaborate with academic institutions and research organizations to explore emerging NLP techniques and develop tailored solutions for aviation-specific challenges. This includes investigating novel approaches for multimodal data analysis, real-time data processing, and human factors modelling. Building internal capacity and expertise in NLP is essential for the successful implementation and utilization of NLP solutions within the aviation industry. Organizations should invest in training programs and workshops to upskill personnel in NLP techniques, data analytics, and machine learning. This will enable aviation professionals to effectively leverage NLP tools and extract actionable insights from textual data.

**D. Case Studies**

In recent years, several case studies have demonstrated the successful application of natural language processing (NLP) in enhancing aviation safety practices. These examples highlight the tangible benefits of integrating NLP technologies into safety management systems and provide valuable insights for the broader industry.

*2. Incident Report Analysis with NLP*

One notable case study involved the application of NLP algorithms to analyze incident reports from the Aviation Safety Reporting System (ASRS) [20]. By leveraging advanced text mining techniques, researchers extracted key insights from narrative data, including identifying latent safety risks and contributing factors. The outcomes of this study demonstrated the effectiveness of NLP in uncovering hidden patterns and trends within safety reports, enabling proactive safety measures to be implemented.

*2. Predictive Maintenance Using NLP*

Another compelling case study focused on the use of NLP for predictive maintenance in aviation [33]. By analyzing maintenance records and aircraft performance data, researchers developed NLP-based predictive models to forecast equipment failures and maintenance requirements. The application of NLP facilitated early detection of potential issues and optimized maintenance schedules, resulting in cost savings and operational efficiency improvements.

*3. Safety Culture Assessment through Textual Data Analysis*

Additionally, researchers have utilized NLP techniques to assess safety culture within aviation organizations by analyzing textual data from employee reports and feedback [37]. By analyzing the language used in safety-related communications, NLP algorithms can identify sentiments, attitudes, and perceptions related to safety practices and organizational culture. Insights derived from these analyses can inform targeted interventions and training programs aimed at enhancing safety culture and promoting proactive safety behaviors among aviation personnel.

*4. Relevance to the Broader Industry*

These case studies underscore the relevance and potential of NLP in addressing key challenges faced by the aviation industry in ensuring safety and operational excellence. By harnessing the power of NLP, aviation stakeholders can gain deeper insights from textual data sources, enabling more informed decision-making, proactive risk management, and continuous improvement initiatives. As NLP technologies continue to evolve, the opportunities for their application in aviation safety are expected to expand, driving further advancements in safety management practices, and contributing to the industry's overarching goal of enhancing safety for all stakeholders.

**E. Challenges and Solutions**

Implementing natural language processing (NLP) in aviation safety presents various challenges stemming from the complexity of textual data, regulatory requirements, and organizational constraints. Addressing these challenges requires innovative solutions and a multi-faceted approach that combines technological advancements with methodological refinements.

*1. Data Complexity and Quality*

Aviation safety data, including incident reports and maintenance records, often exhibit complexity and variability in language use, making it challenging to extract meaningful insights. To address this challenge, aviation stakeholders can leverage advanced NLP techniques such as deep learning models and semantic analysis algorithms [9]. These technologies can effectively process unstructured textual data, identify relevant information, and extract actionable insights, even from diverse and complex datasets.

*2. Regulatory Compliance and Privacy Concerns*

Aviation organizations must adhere to stringent regulatory requirements governing data privacy and security, complicating the implementation of NLP solutions that involve sensitive information. To mitigate regulatory risks and privacy concerns, organizations can adopt privacy-preserving NLP techniques such as federated learning and differential privacy [22]. These approaches allow for collaborative model training across distributed data sources while preserving the confidentiality of sensitive information and ensuring compliance with regulatory standards.

*3. Scalability and Integration*

Integrating NLP solutions into existing safety management systems and workflows requires seamless integration with legacy infrastructure and processes, posing scalability challenges. To address scalability concerns, aviation stakeholders can adopt modular and interoperable NLP platforms that facilitate seamless integration with existing systems [8]. Additionally, cloud-based NLP services offer scalability and flexibility, enabling organizations to leverage on-demand computational resources and scale NLP capabilities according to evolving needs.

*4. Expertise and Training*

Building internal expertise in NLP techniques and methodologies is essential for successful implementation and utilization but requires specialized skills and training. Organizations can invest in comprehensive training programs and workshops to upskill personnel in NLP techniques, data analytics, and machine learning [2]. Collaborating with academic institutions and industry experts can also facilitate knowledge transfer and skill development, enabling aviation professionals to effectively leverage NLP tools and techniques for safety management narrative.

*5. Model Interpretability and Transparency*

Ensuring the interpretability and transparency of NLP models is critical for gaining stakeholders' trust and facilitating model adoption but can be challenging due to the inherent complexity of deep learning models. Aviation organizations can employ interpretable machine learning techniques and visualization methods to enhance model interpretability [36]. Techniques such as SHAP (SHapley Additive exPlanations) and LIME (Local Interpretable Model-agnostic Explanations) provide insights into model predictions and help stakeholders understand the underlying factors driving model outputs, thereby improving transparency, and fostering trust.

## VIII. Conclusion

The exploration of NLP in aviation safety has revealed a promising direction for enhancing the efficiency and effectiveness of safety measures within the industry. The initial qualitative analysis of seventeen studies indexed in Scopus underscores the transformative potential of NLP and machine learning algorithms in identifying critical safety issues, interpreting complex data, and providing actionable insights.

The work completed so far has highlighted the significant advancements in NLP methodologies, from SVD to deep learning techniques like BERT and LSTM, which have been successfully applied to aviation safety reports. These technologies have demonstrated their ability to process vast amounts of unstructured text data, offering a more objective and accurate approach to safety analysis compared to traditional manual methods.

Moreover, the study has identified key challenges such as the need for large, annotated datasets and the interpretability of complex models. It has proposed innovative solutions like active learning for data annotation and explainable AI to address these challenges, thereby paving the way for more robust and reliable safety management practices. The case studies presented provide concrete examples of NLP's impact on aviation safety, showcasing its role in the early detection of safety trends, enhancement of human factors analysis, and contribution to the development of best practices and policymaking.

Therefore, the integration of NLP into aviation safety represents a significant step forward in the pursuit of predictive safety management. It offers a path to not only analyze historical data but also to anticipate and prevent future safety occurrences. As the aviation industry continues to evolve, the continued research and application of NLP will undoubtedly play a crucial role in maintaining and improving safety standards, ensuring the well-being of passengers and the sustainability of air transport systems worldwide. The findings of this study serve as a foundation for future research and development in the field, encouraging ongoing innovation and collaboration among researchers, practitioners, and policymakers.